\documentclass[numsec,webpdf,modern,large]{oup-authoring-template}
\usepackage{csquotes}
\usepackage{pifont}% http://ctan.org/pkg/pifont
\usepackage{amsmath,amssymb}
\usepackage{upgreek}
\usepackage[frozencache,cachedir=minted-cache]{minted}
\usepackage{makecell}
\usepackage{colortbl}
\usepackage{booktabs}
\usepackage{graphicx}
\usepackage{subcaption}
\hypersetup{
	colorlinks = true,
	citecolor = blue
}

\setcitestyle{round}
\definecolor{myred}{RGB}{220, 50, 32}
\definecolor{myblue}{RGB}{0, 90, 181}
\newcommand{\cmark}{\ding{51}}
\newcommand{\xmark}{\ding{55}}

\graphicspath{{Fig/}}

\theoremstyle{thmstyleone}%
\theoremstyle{thmstyletwo}%
\theoremstyle{thmstylethree}%

\begin{document}

\journaltitle{Journal Title Here}
\DOI{DOI HERE}
\copyrightyear{2022}
\pubyear{2019}
\access{Advance Access Publication Date: Day Month Year}
\appnotes{Paper}

\firstpage{1}

\subtitle{Data and text mining}

% \title[A biomedical entity linking benchmark]{A biomedical entity linking benchmark: Rule-based Systems 5 - 0 Pre-trained Language Models}
\title[BELB: a Biomedical Entity Linking Benchmark]{BELB: a Biomedical Entity Linking Benchmark}

\author[1,$\ast$]{Samuele Garda}
\author[2]{Leon Weber-Genzel}
\author[1]{Robert Martin}
\author[1,$\ast$]{Ulf Leser}

\authormark{Garda et al.}

\address[1]{\orgdiv{Computer Science}, \orgname{Humboldt-Universit\"at zu Berlin}, \orgaddress{\street{Rudower Chaussee 25}, \postcode{12489}, \state{Berlin}, \country{Germany}}}
\address[2]{\orgdiv{Center for Information and Language Processing}, \orgname{Ludwig-Maximilians-Universität München}, \orgaddress{\street{Geschwister-Scholl-Platz 1}, \postcode{80539}, \state{München}, \country{Germany}}}

\corresp[$\ast$]{Corresponding author. \href{email:gardasam@informatik.hu-berlin.de}{gardasam@informatik.hu-berlin.de}.\\}
\corresp[$\ast$]{Corresponding author. \href{email:leser@informatik.hu-berlin.de}{leser@informatik.hu-berlin.de}}

\received{Date}{0}{Year}
\revised{Date}{0}{Year}
\accepted{Date}{0}{Year}

%\editor{Associate Editor: Name}

\abstract{
	\textbf{Motivation:} Biomedical entity linking (BEL) is the task of grounding entity mentions to a knowledge base.
	It plays a vital role in information extraction pipelines for the life sciences literature.
	We review recent work in the field and find that,
	as the task is absent from existing benchmarks for biomedical text mining,
	different studies adopt different experimental setups
	making comparisons based on published numbers problematic.
	Furthermore, neural systems are tested
	primarily on instances linked to the broad coverage knowledge base UMLS,
	leaving their performance to more specialized ones, e.g. genes or variants, understudied. \\
	\textbf{Results:} We therefore developed \textbf{BELB},
	a \textbf{B}iomedical \textbf{E}ntity \textbf{L}inking \textbf{B}enchmark,
	providing access in a unified format to 11 corpora linked to 7 knowledge bases
	and spanning six entity types: gene, disease, chemical, species, cell line and variant.
	BELB greatly reduces preprocessing overhead in testing BEL systems on multiple corpora
	offering a standardized testbed for reproducible experiments.
	Using BELB we perform an extensive evaluation of six rule-based entity-specific systems
	and three recent neural approaches leveraging pre-trained language models.
	Our results reveal a mixed picture showing that
	neural approaches fail to perform consistently across entity types,
	highlighting the need of further studies towards entity-agnostic models. \\
	\textbf{Availability:} The source code of BELB is available at: \url{https://github.com/sg-wbi/belb}.
	The code to reproduce our experiments can be found at: \url{https://github.com/sg-wbi/belb-exp}.
	% \textbf{Contact:} \href{name@email.com}{name@email.com}\\
	% \textbf{Supplementary information:} Supplementary data are available at \textit{Journal Name} online.
}

% \boxedtext{
% \begin{itemize}
% \item Key boxed text here.
% \item Key boxed text here.
% \item Key boxed text here.
% \end{itemize}}

\maketitle

\section{Introduction}

The task of assigning entity mentions found in biomedical text to a knowledge base (KB) entity
is known as Biomedical Entity Linking\footnote{
	In some studies \enquote{entity linking} denotes
	the joint entity recognition and linking process.
	We however refer exclusively to the grounding step.
	The task is also known as Named Entity Normalization and we will use
	\enquote{linking}, \enquote{grounding} and \enquote{normalizing}
	interchangeably throughout the text.} (BEL).
Texts in the biomedical domain are rich in ambiguous expressions,
with abbreviation being a prominent example, e.g.:
\enquote{WSS} can be either \enquote{Wrinkly skin syndrome} or \enquote{Weaver-Smith syndrome}.
BEL resolves such ambiguities and is therefore a crucial component in many downstream applications.
For instance it is used to index PubMed \citep{morknlm}, a primary archive of biomedical literature.

\begin{figure*}
	\centering
	\includegraphics[scale=0.25]{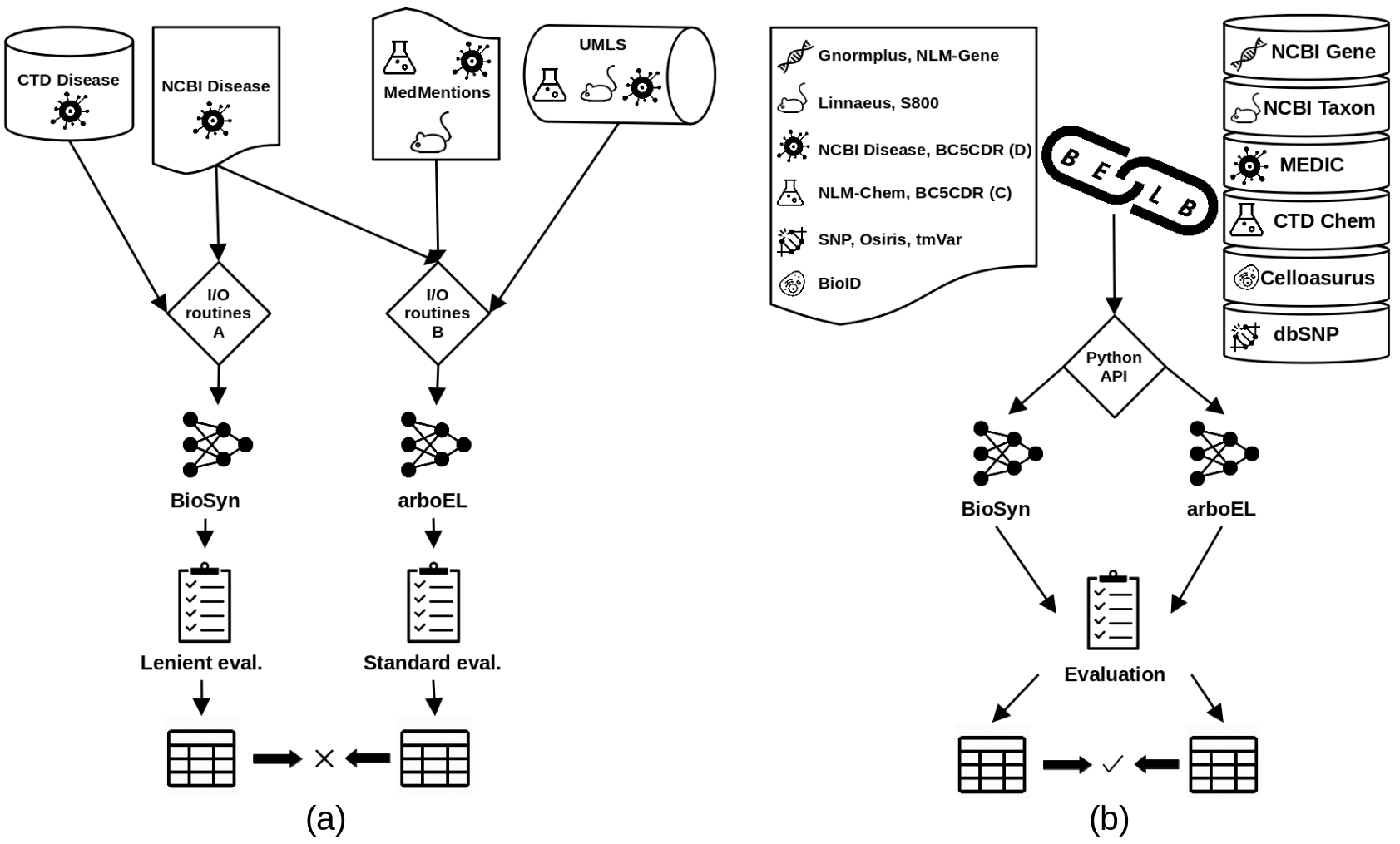}
	\caption{We illustrate the main advantages of BELB.
		In (a) we see the current stand of experimental setups for biomedical entity linking.
		Different studies use different
		(i) preprocessing (I/O routines),
		(ii) combinations of corpora and KBs
		and (iii) evaluation protocols,
		ultimately making published numbers	not directly comparable.
		With BELB (b) researcher have access to (i) uniformly preprocessed corpora and KBs,
		which can be accessed programmatically and (ii) a standardized evaluation protocol
		greatly reducing preprocessing overhead and maximizing comparability and reproducibility.
	}
	\label{fig:belb}
\end{figure*}

Although several benchmarks have been developed for biomedical text mining,
e.g. BLUE \citep{Peng2019} and BLURB \citep{Gu2021},
BEL is notably absent from all of them.
GeneTuring \citep{GeneturingTestHouW2023} contains a module to test normalization,
but covers only genes and is specific for models built on the GPT-3 architecture.
The lack of a standardized evaluation setup translates into a wide variety of approaches:
different studies use different combinations of corpus and KB
and different evaluation protocols.
These differences limit severely direct comparison of results (see Appendix \ref{app:studies_overview}).

In the biomedical domain different entity types
require normalization to different specialized KBs \citep{Wei2019a},
e.g. species to \textsc{NCBI Taxonomy} \citep{TheNcbiTaxonoScott2012}
but genes to \textsc{NCBI Gene} \citep{Brown2015}.
% and diseases to \textsc{CTD Diseases} \citep{ComparativeToxDavis2023}.
Yet, important types such as genes and variants are completely absent
from corpora commonly used to evaluate neural BEL approaches (see \ref{sec:analysis:umls}),
which instead only target UMLS.
Although adapting neural approaches to other KBs is possible,
it leaves open the question of whether their performance transfers across entity types.
Additionally, as corpora are distributed in different formats,
developing new BEL approaches (or adapting existing ones to new corpora)
requires writing new input-output and quality assurance routines,
e.g. to correct wrong mentions boundaries,
increasing the overall engineering turnaround.
% Secondly, and critically, corpora are linked to different KBs (in different versions)
%  (making comparison difficult).

% we first outline its main properties (Section \ref{sec:analysis})
% and use these to derive \textbf{BELB},
To facilitate research on BEL, we introduce \textbf{BELB},
a \textbf{B}iomedical \textbf{E}ntity \textbf{L}inking \textbf{B}enchmark.
BELB provides access to 11 corpora linked to 7 KBs.
All components undergo extensive data cleaning and are converted in a \textit{unified format},
covering six biomedical entities (gene, disease, chemical, species, cell lines and variants).
As show in Figure \ref{fig:belb}, BELB significantly lowers the barrier for research in the field,
allowing to (i) train models on corpora in the highest quality possible and
(ii) fairly compare them against other approaches with minimal preprocessing overhead
(see Appendix \ref{sec:showcase} for a simple showcase).
Using BELB, we perform an extensive comparison of \textbf{six rule-based domain-specific systems
	and three neural methods}.
Our findings show that results of neural approaches do not transfer across entity types,
with specialized rule-based systems still being the best option for the gene and disease entity types.
We hope that our publicly available benchmark will be adopted by future work
allowing to fairly evaluate approaches and accelerate progress
towards more robust neural models.

% With the advent of pre-trained language models tailored,
% there have been extensive efforts to create benchmarks to test their capabilities.
% BLUE (Biomedical Language Understanding Evaluation) is a benchmark consisting of 10 corpora encompassing 5 different tasks \citep{},
% which was extended by BLURB (Biomedical Language Understanding and Reasoning Benchmark) to include 13 corpora and 7 tasks \citep{}.
% However, none of them offers integration with a KB,
% which is an essential component to streamline training and evaluation.
% While all resources reported offer access to corpora
% which in principle could support the linking task,
% only GeneTuring specifically includes the task in their benchmarking

% The two KBs differ considerably in the number of entities and
% as using a smaller KB alone can improve results \citep{Wu2020}
% Recent studies present as well differences in the of evalution protocol
% further impacting comparison in reported performance
% as shown in Table \ref{tab:approaches},
% where we report a summary of recent approaches,
% , which puts a strong emphasis on some entity types
% This can be problematic as UMLS 
% and almost completely ignores others such as variants .
% As a result, rule-based entity-specific systems
% are still the de facto standard \citep{Wei2019a, Bern2AnAdvanMujeen} in the field.

\section{Materials and methods}~\label{sec:belb}

\begin{table*}[!h]
	\centering
	\resizebox{\textwidth}{!}{
		\begin{tabular}{l|c|c|c|c}
			                                                                & Documents (train / dev / test) & Annotations (train / dev / test) & 0-shot          & Stratified      \\
			\toprule
			\textbf{Disease}                                                &                                &                                  &                 &                 \\
			\quad \texttt{NCBI Disease} \citep{Dogan2014}                   & 592 / 100 / 100                & 5,133 / 787 / 960                & 150 (15.62\%)   & 185 (19.27\%)   \\
			\quad \texttt{BC5CDR} (D) \citep{Li2016a}                       & 500 / 500 / 500                & 4,149 / 4,228 / 4,363            & 388 (8.89\%)    & 765 (17.53\%)   \\
			\hline
			\textbf{Chemical}                                               &                                &                                  &                 &                 \\
			\quad \texttt{BC5CDR} (C)\citep{Li2016a}                        & 500 / 500 / 500                & 5,148 / 5,298 / 5334             & 1,038 (19.46\%) & 415 (7.78\%)    \\
			\quad \texttt{NLM-Chem} $\dagger$ \citep{NlmChemBc7MIslama2022} & 80 / 20 / 50                   & 20,796 / 5,234 / 11514           & 3,908 (33.94\%) & 1,534 (13.32\%) \\
			\hline
			\textbf{Cell line}                                              &                                &                                  &                 &                 \\
			\quad \texttt{BioID} $\ddag$ \citep{arighi2017bio}              & 231 / 59 / 60                  & 3,815 / 1,096 / 864              & 158 (18.29\%)   & 45 (5.21\%)     \\
			\hline
			\textbf{Species}                                                &                                &                                  &                 &                 \\
			\quad \texttt{Linnaeus} $\dagger$ \citep{LinnaeusASpeMartin}    & 47 / 17 / 31                   & 2,115 / 705 / 1,430              & 385 (26.92\%)   & 58 (4.06\%)     \\
			\quad \texttt{S800} \citep{Pafilis2013}                         & 437 / 63 / 125                 & 2,557 / 384 / 767                & 363 (47.33\%)   & 107 (13.95\%)   \\
			\hline
			\textbf{Gene}                                                   &                                &                                  &                 &                 \\
			\quad \texttt{GNormPlus} \citep{Wei2015}                        & 279 / 137 / 254                & 3,015 / 1,203 / 3,222            & 2,822 (87.59\%) & 163 (5.06\%)    \\
			\quad \texttt{NLM-Gene} \citep{Islamaj2021}                     & 400 / 50 / 100                 & 11,263 / 1,371 / 2,729           & 1,215 (44.52\%) & 353 (12.94\%)   \\
			\hline
			\textbf{Variant}                                                &                                &                                  &                 &                 \\
			\quad \texttt{SNP} \citep{ChallengesInTThomas2011}              & - / - / 292                    & - / - / 517                      & -               & -               \\
			\quad \texttt{Osiris v1.2} \citep{Osirisv12ANFurlon2008}        & - / - / 57                     & - / - / 261                      & -               & -               \\
			\quad \texttt{tmVar v3} \citep{Wei2022}                         & - / - / 214                    & - / - / 1,018                    & -               & -               \\
			\midrule
			\textbf{UMLS}                                                   &                                &                                  &                 &                 \\
			\quad \texttt{MedMentions} \citep{mohanmedmentions}             & 2,635 / 878 / 879              & 122,178 / 40,864 / 40,143        & 8,167 (20.34\%) & 7,945 (19.79\%) \\
			\bottomrule
		\end{tabular}
	}
	\caption{Overview of the corpora available in BELB with their primary characteristics: number of documents, annotations
		and how many of them are zero-shot (unseen entities) or stratified (seen entity but unseen name).
		$\ddag$ Full text
		$\ddag$ Figure captions}~\label{tab:corpora:overview}
\end{table*}

\begin{table*}[!h]
	\centering
	\resizebox{\textwidth}{!}{
		\begin{tabular}{l|c|c|l|l|l|l}
			                                                             & Version              & History & Entities      & Names         & Synonyms & Homonyms (PN)           \\%                & SFS (0,1) $\uparrow$
			  \toprule                                                                                                                                                             %                       
			\textbf{Disease}                                             &                      &         &               &               &          &                         \\%                                                           &
			\quad \textsc{CTD Diseases} \citep{ComparativeToxDavis2023}  & monthly $\dagger$    & \xmark  & 13,188        & 88,548        & 6.71     & 0.39\% (-)              \\%                 & 0.54
			  \hline                                                                                                                                                               %                       
			\textbf{Chemical}                                            &                      &         &               &               &          &                         \\%                                                            &
			\quad \textsc{CTD Chemicals} \citep{ComparativeToxDavis2023} & monthly  $\dagger$   & \xmark  & 175,663       & 451,410       & 2.56     & - (-)                   \\%                  & 0.61
			  \hline                                                                                                                                                               %                       
			\textbf{Cell line}                                           &                      &         &               &               &          &                         \\%                                                           &
			\quad \textsc{Cellosaurus} \citep{TheCellosaurusBairoc2018}  & -                    & \cmark  & 144,568       & 251,747       & 1.74     & 3.21\% (1.22\%)         \\%                & 0.76
			  \hline                                                                                                                                                               %                       
			\textbf{Species}                                             &                      &         &               &               &          &                         \\%                                                           &
			\quad \textsc{NCBI Taxonomy} \citep{TheNcbiTaxonoScott2012}  & -                    & \cmark  & 2,491,364     & 3,783,882     & 1.51     & 0.04\% (-)              \\%             & 0.90
			  \hline                                                                                                                                                               %                       
			\textbf{Gene}                                                &                      &         &               &               &          &                         \\%                                          &
			\quad \textsc{NCBI Gene} \citep{Brown2015}                   & -                    & \cmark  & 42,252,923    & 105,570,090   & 2.49     & 47.37\% (8.32\%)        \\%                                & 0.20
			\qquad \texttt{GNormPlus} subset                             &                      &         & 703,858       & 2,455,772     & 3.48     & 50.79\% (9.13\%)        \\%                               & 0.26
			\qquad \texttt{NLM-Gene} subset                              &                      &         & 873,015       & 2,913,456     & 3.33     & 53.61\% (9.55\%)        \\%                & 0.26
			  \hline                                                                                                                                                               %                       
			\textbf{Variant}                                             &                      &         &               &               &          &                         \\%                                       &
			\quad \textsc{dbSNP} \citep{Sherry2001}                      & build 156  $\dagger$ & \cmark  & 1,053,854,063 & 3,119,027,235 & 2.95     & 1,557,105,418 (49.92\%) \\%             & 0.65
			  \midrule                                                                                                                                                             %                       
			\textbf{UMLS}                                                &                      &         &               &               &          &                         \\%                                           &
			\quad \textsc{UMLS} \citep{Bodenreider2004}                  & 2017AA (full)        & -       & 3,464,809     & 7,938,833     & 2.29     & 2.07\% (0.16\%)         \\% & 0.82
			  \bottomrule
		\end{tabular}
	}
	\caption{Overview of the KBs available in BELB according to their entity type.
		We report the number of entities, synonyms per entity, homonyms and how many of them are the primary name (PN).
		$\dagger$ No archive of previous versions is provided
	}~\label{tab:kbs:overview}
\end{table*}

In this study we introduce BELB,
a benchmark for standardized evaluation of models for BEL.
The task is formulated as predicting an entity $e \in E$ from a KB
given a document $d$ and an entity mention $m$,
a pair of start and end positions $\langle m_{s}, m_{e} \rangle$ indicating a span in $d$.
We use BELB to compare rule-based domain-specific systems
and state-of-the-art neural approaches.
In all experiment we use in-KB gold mentions:
each mention has a valid gold KB entity \citep{GerbilBenchmRoder2018}
and its position in $d$ is given.

\subsection{Biomedical Entity Linking Benchmark}

We report an overview of the 11 corpora and 7 KBs available in BELB
in Table \ref{tab:corpora:overview} and \ref{tab:kbs:overview}, respectively.
Their detailed description can be found in Appendix \ref{app:resources}.
In the following we outline crucial properties of BEL and
highlight how they are accounted for in BELB,
allowing it to comprehensively analyze and fairly evaluate BEL models.

\subsubsection{Specialized knowledge bases}\label{sec:analysis:umls}

% For instance, PubTator,
% a system widely adopted in downstream applications \citep{Wei2019a}, supports six.

In biomedical information extraction instances of multiple entity types
are linked to specialized KBs \citep{Wei2019a}.
However, recent studies in the NLP community primarily focus on the \texttt{MedMentions} corpus linking to \textsc{UMLS}
(\citealp{Liu2021a,zhang2022knowledge,Agarwal2022} inter alia).
% which presents significant gaps and area of emphasis \citep{BiomedicalTextRodrig2009}.
Additionally, in \texttt{MedMentions}, entity types such as diseases and genes
are covered only marginally or not at all, respectively
(see Appendix \ref{sec:medmentions}).

This calls into question how well results obtained in this setting
can be transferred for instance to publications in genomics or
molecular biology in general.
In BELB we cover six entity types
(gene, species, disease, chemicals, cell line and variant)
represented by 11 corpora linked to 7 specialized KBs
(for comparison with previous studies we include \textsc{UMLS} as well).
We design a \textit{unified schema} to harmonize all KBs (see Appendix \ref{app:schema}).
This allows to test a model's ability to
preserve performance across multiple pairs of corpus and KB
with minimal preprocessing overhead.

\subsubsection{Unseen entities and synonyms}\label{sec:analysis:zeroshot}
Corpora typically cover only a small fraction of all entities in a KB.
Additionally, biomedical entities present multiple names (\textit{synonyms}),
e.g. both \enquote{Oculootofacial dysplasia}  and \enquote{Burn-Mckeown Syndrome}
are valid names of \enquote{MeSH:C563682}.
Hence if an entity is in the training set,
it does not imply that all its surface forms are included.
In BELB we assign a unique identifier to each mention
and provide lists of mentions of (i) unseen entities, i.e. present in the test set but not in the train one (\textit{zero-shot}) and
(ii) train entities occurring in the test set but with different (case-insesitive) surface forms \citep{Tutubalina2020}.
This allows to easily report a model's performance in
(i) generalizing to new entities and
(ii) recognizing known ones appearing in different forms.

\begin{table*}[!htbp]
	\centering
	\resizebox{15cm}{!}{
		\begin{tabular}{l|l|l}
			   & Example                                                                                                         & Entity                   \\
			\toprule
			a) & Features of ARCL type II overlap with those of \underline{Wrinkly skin syndrome} (\textbf{WSS})                 & MeSH:C536750             \\
			b) & \underline{Weaver-Smith syndrome} (\textbf{WSS}) is a Mendelian disorder of the epigenetic machinery            & MeSH:C536687             \\
			\midrule
			c) & $\boldsymbol{\upalpha}$\textbf{2microglobulin} exacerbates brain damage after stroke in \underline{rats}.       & \textsc{NCBI Gene}:24153 \\
			d) & The \underline{T67 cell line} produced the proteinase inhibitor $\boldsymbol{\upalpha}$\textbf{2microglobulin}. & \textsc{NCBI Gene}:2     \\
			e) & We identified the novel mutation \textbf{c.908G$>$A} within exon 8 of the \underline{CTSK} gene.                & rs756250449              \\
			f) & The patient was compound heterozygous of the \textbf{c.908G$>$A} mutation in the \underline{SLC17A5} gene.      & rs1057516601             \\
			g) & The GSK650394 inhibitor is used to suppress \underline{SGK1} expression in \textbf{PC12 cells}.                 & CVCL\_S979               \\
			h) & Effects of topography on \underline{rat} pheochromocytoma cell, \textbf{PC12 cells}, neuritogenesis.            & CVCL\_0481               \\
			\bottomrule
			% 3) & Analysis of the \underline{VWF} gene showed a novel, \textbf{heterozygous T-->G transversion at nucleotide 4508}. \\
			% 4) & We identified a novel \textbf{nine-nucleotide deletion starting at position 1952} in the \underline{PIX} gene.   \\
			 % 5) & Analysis of $\boldsymbol{\upalpha}$\textbf{2-microglobulin} in \underline{PC12 cells} following induction by TPA. \\
		\end{tabular}
	}
	\caption{Example of homonym mentions (in \textbf{bold}) requiring specific contextual information (\underline{underlined}) for linking.
	}~\label{tab:homonyms}
\end{table*}

% This may require as well domain-specific multi-hop reasoning as in example (d),
% where \enquote{T67 cell line} must be recognized to be a \textit{human} cell line.
\subsubsection{Homonyms}\label{sec:analysis:homonyms}
Discriminating mentions with the same surface form but representing different entities
(\textit{homonyms}) by their context is indispensable to BEL.
This is because in biomedical KBs the same synonym can be associated to multiple entities.
This is especially the case of abbreviations. For instance, as in example (a) in Table \ref{tab:homonyms},
\enquote{WSS} is the abbreviated form of two syndromes and it appears twice in \textsc{CTD Diseases}.
Another example are genes present in multiple species, as in (c),
where the string \enquote{rats} is essential for correct normalization,
as \enquote{$\alpha$2microglobulin} could refer either to the human or rat gene.
Indentifying contextual information can be non-trivial, e.g
(c) is the title of a publication,
but the text may describe general characteristics
of \enquote{$\alpha$2microglobulin} introducing textual cues pointing to the human gene.
Additionally, this information is not always explicitly expressed
and may emerge via other patterns.
In example (e) \enquote{PC12} denotes a human cell line,
whereas in (f) it refers to the rat one.
This can be inferred from the capitalized gene mention \enquote{SGK1}
in (e) which conventionally denotes human genes.
By introducing entity types such as genes and variants,
BELB allows to probe a model's ability to (i) identify contextual information
and (ii) handle highly ambiguous search spaces (KB).

\subsubsection{Scale} As mentioned in Section \ref{sec:analysis:umls}
studies on neural methods have primarily targeted UMLS,
which, as shown in Table \ref{tab:kbs:overview},
is one and three order of magnitude(s) smaller than \textsc{NCBI Gene}
and \textsc{dbSNP}, respectively.
With its unified format
BELB allows to easily test how implementations scale to these large KBs.

\subsubsection{Synchronization of KB versions}\label{sec:analysis:synchronization}
Entity linking is dynamic by nature:
over the years entities in KBs are replaced or become obsolete.
For instance, in \texttt{GNormPlus} mentions of \enquote{MDS1} are linked to \textsc{NCBI Gene} entity \enquote{4197},
which was subsequently replaced by \enquote{2122}.
As several KBs do not have a versioning system (see Table \ref{tab:kbs:overview}),
it is often not possible to retrieve the exact KB used to create a corpus.
Failing to account for these changes may introduce a notable amount of noise in measuring performance
of high quality systems.
BELB offers access to the KB \textit{history} if available,
i.e. a table tracking all changes in the entities.
In our preprocessing we update all corpora with the KB version at hand
and remove mentions linked to obsolete entities.
This allows as well to update the predictions of
systems shipping with a pre-processed (i.e. non-trivially swappable) KB
on corpora created after their release, allowing for fair comparison \textit{over time}.

\subsection{Evaluated approaches}~\label{sec:analysis:eval}

We use BELB to perform an extensive evaluation of rule-based
and neural methods.
We now present the selected approaches for evaluation.
We stress that we do not re-implement any method (we rely on the original code).

\subsubsection{Rule-based entity-specific systems}

We compare the performance of neural models linking to KBs
for which specialized systems have been developed,
as these are still the de facto standard in BEL \citep{Wei2019a, Bern2AnAdvanMujeen}.
Specifically we test the following rule-based methods:
\textbf{GNormPlus} \citep{Wei2015} for genes (\textsc{NCBI Gene}) ,
\textbf{SR4GN} \citep{Wei2012} for species (\textsc{NCBI Taxonomy})
and \textbf{tmVar v3} \citep{Wei2022} for variants (\textsc{dbSNP}).
For \textsc{UMLS} we employ \textbf{SciSpacy} \citep{Neumann2019}.
We label them rule-based entity-specific (RBES) as
for linking they rely on a mixture of string matching approaches
and ad-hoc rules tailored to a specific entity type. For diseases and chemicals,
we include in the RBES category two systems which are only \textit{partly} rule-based
(stretching our definition),
as they better represent the state-of-the-art of disease/chemical-specific models.
We use \textbf{TaggerOne} \citep{Leaman2016}, a semi-Markov model, for diseases,
and opt for the system that won the BioCreative VII NLM-Chem track \citep{ChemicalIdentiAlmeid2022}
for chemicals (\textbf{BC7T2W}), which uses both string matching and neural embeddings.
% preferring it over MetaMap \citep{aronson1994exploiting}
% as it allows to run recognition and normalization separately (see Section \ref{sec:disentangling_ner_nen}).
To the best of our knowledge there exists no linking approach specific for cell lines.
We therefore use a fuzzy string matching approach based on Levenshtein distance (\textbf{FuzzySearch}).
For detailed descriptions and information on specific implementations we refer the reader to Appendix \ref{app:systems:rb}.
All of these systems do not require re-training as either
(i) their normalization component is completely rule-based (SR4GN, tmVar, SciSpacy) or
(ii) models trained on the BELB corpora are provided along with the code (GNormPlus, TaggerOne, BC7T2W).

\subsubsection{Neural systems}
We train the following neural models on the train split of each BELB corpus
(see Appendix \ref{app:systems:plm} for training details).

\noindent \textbf{BioSyn} \citep{Sung2020} is a dual encoder architecture.
% initialized with BioBERT \citep{lee2020biobert} weights.
Importantly, BioSyn does not account for context, i.e. it uses only entity mentions.
The model is trained via \enquote{synonym marginalization}:
it learns to maximize the similarity (inner product)
between a mention embedding and all the synonyms embeddings of the gold entity.
At inference it retrieves the synonyms most similar to the given test mention,
i.e. it relies on a lookup from synonym to entity.
We prefer BioSyn over SapBERT \citep{Liu2021a} as the latter is primarily a pre-training strategy.

% As enitity mentions are annotated with atomic labels (see Section \ref{sec:analysis:entities}), 
% for training it necessary to determine the synonym to be generated.
% The authors propose to select the one with the highest cosine similarity with the mention
% (all represented as as 3-gram TF-IDF vectors). 
\noindent \textbf{GenBioEL} \citep{GenerativeBiomYuan} is an encoder-decoder model.
% based on a fine-tuned BART model \citep{BartDenoisingLewis2020}. 
As input it takes a text with a single mention marked with special tokens.
The system is then trained to generate a synonym.
At inference it ensures that the prediction is a valid KB synonym
by constraining the generation process with a prefix-tree created from all KB synonyms.
Similar to BioSyn, this approach represents KB entities by their synonyms.
The authors propose as well \enquote{KB-Guided Pre-training},
i.e. a method based on prompting to pre-train GenBioEL on the KB,
which we however do not deploy.
This is because  (i) it would introduce an advantage over other neural methods and
(ii) it is too computationally expensive to run for each KB.
% \footnote{Authors report to use six A100 GPUs to pre-train on a subset of UMLS.}.

\noindent \textbf{arboEL} \citep{Agarwal2022} is a dual encoder as well.
The authors propose to construct k-nearest neighbor graphs over mention and entity clusters.
Using a pruning algorithm they then generate directed minimum spanning trees rooted at entity nodes,
and use the edges as positive examples for training.
At inference they use the entity present in the mention's cluster.
Notably, arboEL learns \textit{entity embeddings}, encoded as a concatenated list of synonyms.
The authors use as well a cross-encoder \citep{Wu2020}, i.e.
using the top-64 entities retrieved by a trained arboEL
as hard negatives (training) and as linking candidates (inference)
for a second reranking model.
In our experiments we do not make use of this extension as
it is not strictly part of the arboEL algorithm.
% i.e. it can in principle be used by any model.

\subsection{Evaluation protocol}\label{sec:metrics}

We now describe in detail the evaluation protocol which we followed in our experiments.
For all systems we report the \textit{mention-level} recall@1 (accuracy),
since RBES approaches generate only a single candidate.
% As one of the primary use cases for entity linking is document indexing \citep{ChemicalIdentiLeaman2023},
% we compute as well the \textit{document-level} precision, recall and F1-score.

\subsubsection{Synonym as entity proxy}~\label{sec:syn_as_ent}
Approaches using strings as proxies for entities (BioSyn, GenBioEL) cannot meaningfully resolve ambiguous mentions.
That is, for a mention of rat \enquote{$\alpha$2microglobulin},
they would return a list containing \textit{both} \textsc{NCBI Gene} \enquote{2} (human) and
\enquote{24153} (rat). \citet{Sung2020} introduced a \textit{lenient} evaluation,
which considers a prediction correct if any of the returned entities matches the gold one.
As reported by \citet{zhang2022knowledge}, this largely overestimates performance.
Following their suggestion, we opt for a \textit{standard} evaluation,
which randomly samples one prediction from the returned list.
However, as one of the aims of BEL is direct deployment in extraction pipelines,
e.g. for constructing gene networks \citep{AssemblyOfACLehman2015},
we also include a \textit{strict} evaluation in which all such cases, i.e. multiple predictions,
are considered incorrect.

\subsubsection{Disentangling recognition and linking}~\label{sec:disentangling_ner_nen}
Some RBES systems (GNormPlus, SR4GN, TaggerOne, tmVar)
perform entity recognition and linking jointly (see Appendix \ref{app:systems:rb}).
Due to false negatives in the NER step we cannot obtain their results on the full test set.
To ensure that we are measuring the performance on the same instances for all methods,
for corpora whose reference RBES system is a joint approach,
we use only the test mentions which are correctly identified during entity recognition.
For instance, for \texttt{NLM-Gene} we use only 73\% of the test mentions,
i.e. those correctly recognized by GNormPlus (see Table \ref{tab:ner_recall} for other corpora).
As correct recognition correlates with correct normalization,
our evaluation protocol probably introduces a bias towards RBES systems (see Section \ref{sec:discussion}).

\subsubsection{Multiple gold entities}~\label{sec:multiple_gold}
Mentions in biomedical corpora can provide multiple normalizations.
Common instances are \textit{composite mentions}, e.g. \enquote{breast and squamous cell neoplasms}
and ambiguous ones, e.g. \enquote{Toll-like receptor} (\enquote{Toll-like receptor 2}, \enquote{4} and \enquote{9}).
Whether these cases are logical AND or OR is not always specified in the annotation guidelines.
We opt for the more lenient OR interpretation
and consider a prediction correct if it matches one of the gold entities.

\section{Results}~\label{sec:results}

\begin{table}[!h]
	\centering
	\resizebox{\columnwidth}{!}{
		\begin{tabular}{l|l|l|l|l}
			Entity type                 & \multicolumn{3}{c}{}                              \\
			\quad Corpus                & RBES                 & BioSyn & GenBioEL & arboEL \\
			\toprule
			\textbf{Disease}            & 0.94                 & 0.87   & 0.90     & 0.87   \\
			\quad \texttt{NCBI Disease} & 0.94                 & 0.83   & 0.89     & 0.86   \\
			\quad \texttt{BC5CDR (D)}   & 0.94                 & 0.88   & 0.92     & 0.88   \\
			\hline
			\textbf{Chemical}           & 0.72                 & 0.72   & 0.81     & 0.77   \\
			\quad \texttt{BC5CDR (C)}   & 0.82                 & 0.85   & 0.95     & 0.88   \\
			\quad \texttt{NLM-Chem}     & 0.67                 & 0.67   & 0.75     & 0.72   \\
			\hline
			\textbf{Cell line}          &                      &        &          &        \\
			\quad \texttt{BioID}        & 0.82                 & 0.82   & 0.96     & 0.95   \\
			\hline
			\textbf{Species}            & 0.97                 & 0.91   & 0.85     & 0.76   \\
			\quad \texttt{Linnaeus}     & 0.99                 & 0.93   & 0.81     & 0.74   \\
			\quad \texttt{S800}         & 0.93                 & 0.88   & 0.93     & 0.79   \\
			\hline
			\textbf{Gene}               & 0.82                 & -      & 0.17     & 0.30   \\
			\quad \texttt{GNormPlus}    & 0.87                 & OOM    & 0.21     & 0.36   \\
			\quad \texttt{NLM-Gene}     & 0.76                 & OOM    & 0.13     & 0.23   \\
			\hline
			\textbf{Variant}            & 0.91                 & -      & -        & -      \\
			\quad \texttt{SNP}$\dagger$ & 0.94                 & -      & -        & -      \\
			\quad \texttt{Osiris v1.2}  & 0.91                 & -      & -        & -      \\
			\quad \texttt{tmVar v3}     & 0.88                 & -      & -        & -      \\
			\midrule
			\textbf{UMLS}               &                      &        &          &        \\
			\quad \texttt{MedMentions}  & 0.58                 & OOM    & 0.57     & 0.69   \\
		\end{tabular}
	}
	\caption{Performance of all baselines on BELB (test set). All scores are \textit{mention-level} recall@1.
		OOM: out-of-memory ($>$200GB)
	}~\label{tab:results:mention}
\end{table}

Table \ref{tab:results:mention} reports the results of neural models
and entity-specific models grouped under the RBES category.
For results with strict evaluation (Section \ref{sec:syn_as_ent}) and on the full test sets (Section \ref{sec:disentangling_ner_nen})
please see Table \ref{tab:results:mention_strict} and \ref{tab:results:full}, respectively.
We observe that performance of neural models varies significantly across entity types,
with disease and genes corpora incurring the most significant drop.

\begin{table}[!h]
	\centering
	% \resizebox{\columnwidth}{!}{
	\begin{tabular}{l|l|l}
		                      & \texttt{NCBI Disease} & \texttt{GNormPlus} \\
		\toprule
		RBES                  & 0.94                  & 0.87               \\
		\midrule
		BioSyn                & 0.83                  & OOM                \\
		\quad + abbr. res.    & 0.88 (+0.5\%)         & -                  \\
		\quad + lenient eval. & 0.88 (+0.5\%)         & -                  \\
		\hline
		GenBioEL              & 0.89                  & 0.21               \\
		\quad + abbr. res.    & 0.91 (+0.2\%)         & 0.20 (-0.1\%)      \\
		\quad + lenient eval. & 0.90 (+0.1\%)         & 0.86 (+65\%)       \\
		\hline
		arboEL                & 0.86                  & 0.36               \\
		\quad + abbr. res.    & 0.86 (-)              & 0.36 (-)           \\
		\quad + lenient eval. & -                     & -                  \\
		 % \bottomrule
	\end{tabular}
	% }
	\caption{Relative improvement of neural models with resolved abbreviations
		and a lenient evaluation in case of multiple \emph{predicted} entities.
		OOM: out-of-memory ($>$200GB)
	}~\label{tab:results:ablations}
\end{table}

\noindent \textbf{Homonyms} Beside the implicit bias towards RBES approaches (see Section \ref{sec:disentangling_ner_nen}),
we hypothesize that an important factor at play are homonyms.
RBES systems use ad-hoc components to handle these challenging cases.
For instance GNormPlus directly integrates Ab3P \citep{AbbreviationDeSohn2008},
a widely adopted abbreviation resolution tool,
and SR4GN, which is specifically developed for cross-species gene normalization.
Neural models lack these components, and synonym-based approaches
are significantly impacted by random selection in case of homonyms.
In table \ref{tab:results:ablations} we show that if we
resolve abbreviation with Ab3P there is a notable improvement
in performance for diseases. Similarly, if we use a lenient evaluation
(see Section \ref{sec:syn_as_ent}), GenBioEL is almost on par with GNormPlus on genes.
In contrast, abbreviation resolution has no impact on arboEL.
We argue that this is due to the fact that arboEL
uses entity embeddings, which benefit less by long forms mentions.
Secondly, as entity embeddings require to learn a compressed
entity representation, arboEL is affected by the limited size
of the corpora. This is supported by results on MedMentions,
which is one order of magnitude larger than other BELB corpora,
where arboEL is confirmed the state-of-the-art approach.

% We hypothesize that neural systems struggle particularly in this setting
% as they need to learn from the minimal training data the salient patterns
% necessary to solve the task.
% In case of arboEL, which truncates the context window to 128 tokens,
% relevant contextual information may be not present at all.
%
\begin{table}[!h]
	\centering
	\begin{tabular}{l|l|l}
		         & Unseen entities & Unseen synonyms \\
		\toprule
		RBES     & 0.65            & 0.40            \\
		GenBioEL & 0.50            & 0.48            \\
		arboEL   & 0.45            & 0.50            \\
	\end{tabular}
	\caption{Results on unseen entities and synonyms (\textit{mention-level} recall@1).}~\label{tab:results:subsets}
\end{table}

\noindent \textbf{Unseen entities and synonyms} In Table \ref{tab:results:subsets}
we see that neural approaches are outperformed by RBES systems on mentions of unseen entities
while we the opposite happens with unseen synonyms of train entities.
This can be explained by the fact that
as string-matching approaches have direct access to the KB
they are better suited for the zero-shot cases.
If training data is available, neural representation are superior instead,
as they can leverage representations learned from context.

\noindent \textbf{Scale} Neural models implementations fail to scale to large KBs
such as \textsc{NCBI Gene} or \textsc{dbSNP}.
In our experiments resorted to use the \textsc{NCBI Gene} subset
determined by the species of the entities found in the gene corpora (see Table \ref{tab:kbs:overview}).
This reflects a common real-world use case scenario,
since often only a specific subset of species is relevant for linking (e.g. human and mouse).
For dbSNP we are not aware of a valid criterion
to subset it and we are therefore unable to run neural models
on variants corpora.

\noindent \textbf{Synchronization of KB version} Corpora are only sparsely
affected by changes in entities.
However, if they are not handled properly,
in \texttt{BC5CDR (C)} and \texttt{Linnaeus} (the most affected corpora in BELB),
even if evaluating a perfect system,
we would register an error rate of 4.56\% and 3.57\% respectively
(see Appendix \ref{app:update_corpora}).

% \begin{tabular}{l|l|l|l|l|l|l}
% 	% ZERO-SHOT
% 	         & gene & disease & chemical & species & cell\_line & umls \\
% 	rbes     & 0.84 & 0.90    & 0.66     & 0.97    & 0.91       & 0.52 \\
% 	genbioel & 0.20 & 0.74    & 0.65     & 0.57    & 0.85       & 0.50 \\
% 	arboel   & 0.27 & 0.46    & 0.52     & 0.45    & 0.77       & 0.47 \\
% 	biosyn   & -    & 0.75    & 0.66     & 0.76    & 0.91       & -    \\
% \end{tabular}
%
% \begin{tabular}{l|l|l|l|l|l|l}
%
% 	% STRATIFIED
% 	         & gene & disease & chemical & species & cell\_line & umls \\
% 	rbes     & 0.83 & 0.72    & 0.45     & 0.82    & 0.24       & 0.34 \\
% 	genbioel & 0.17 & 0.67    & 0.55     & 0.68    & 0.98       & 0.43 \\
% 	arboel   & 0.35 & 0.66    & 0.57     & 0.69    & 1.00       & 0.47 \\
% 	biosyn   & -    & 0.61    & 0.40     & 0.87    & 0.31       & -    \\
% \end{tabular}

\section{Discussion}~\label{sec:discussion}

We strived to include in BELB as many corpora and KBs as possible,
prioritizing those which are most commonly used by the community.
% However, in BELB we focus on a specific setting,
% which we believe already poses a hard challenge.
We leave as further improvement the expansion to other
important research directions
as applications to clinical notes \citep{The2019N2c2ULuoY2020}
and other languages as Spanish \citep{miranda2022overview} and
German \citep{10.1093/jamiaopen/ooab025}.
% to other important (i) corpora and KBs, e.g. CRAFT \citep{cohen2017colorado} and Uniprot \citep{UniprotTheUnNone2023}

Our evaluation showed that neural approaches fail to perform consistently across all BELB instances,
especially on genes, where RBES approaches are still far superior.
However, as reported in Section \ref{sec:disentangling_ner_nen},
our evaluation protocol introduces a bias towards RBES systems
by considering exclusively the test mentions they correctly identify.
Nonetheless, we believe that this is the best approximation
to compare results across all methods.
We note as well the lack of hyperparameter exploration in neural models.
Due to the high computational resources necessary we rely on the default ones reported by the authors.
It is therefore likely that optimizing them may result in better numbers.
Further improvements may possible by pre-training on the KB \citep{Liu2021a,GenerativeBiomYuan}
and refining candidates with a cross-encoder \citep{Agarwal2022}.
RBES systems are advantaged by the use of ad-hoc components to handle homonyms.
In Table \ref{tab:results:ablations} we show that introducing similar approaches for neural models
could significantly improve their performance. However,
as the neural paradigm is based on learning task-related capabilities from data \citep{bengio2013representation},
we believe that future studies should nevertheless continue to investigate entity-agnostic models,
rather than falling back to custom-made hand-crafted heuristics.

\section{Conclusion}

We presented BELB, a benchmark to standardize experimental setups
for biomedical entity linking. We conducted an extensive evaluation
of rule-based entity-specific systems and recent neural approaches.
We find that the first are still the state-of-the-art on entity-types
not explored by neural approaches, namely genes and variants.
We hope that BELB will encourage future
studies to compare approaches with a common testbed
and to address current limitations of neural approaches.

\section*{Acknowledgements}

Samuele Garda and Robert Martin are supported by the \textit{Deutsche Forschungsgemeinschaft} as part of the
research unit \enquote{Beyond the Exome}.

%USE THE BELOW OPTIONS IN CASE YOU NEED AUTHOR YEAR FORMAT.
\bibliographystyle{abbrvnat}
\bibliography{reference}

%%%%%%%%%%%%%%

\begin{appendices}
	\section{Neural biomedical entity linking}\label{app:studies_overview}

	\begin{table*}[!htbp]
		\centering
		\resizebox{\textwidth}{!}{
			% \begin{tabular}{l|c|c|c|l|l|l}
			\begin{tabular}{l|c|c|c}
				Model                                                   & \multicolumn{3}{c}{}                                                                                                                       \\
				\quad Abbr. res. / Pre-train / Lenient eval.            & BC5CDR (Disease / Chemical)                                     & NCBI Disease                                            & MedMentions    \\
				\toprule
				BioSyn \citep{Sung2020}                                 &                                                                 &                                                         &                \\
				\quad \cite{AbbreviationDeSohn2008} / - / yes           & \cellcolor{blue!25}{93.2 (CTD Diseases) / 96.6 (CTD Chemicals)} & \cellcolor{red!25}{91.1 (CTD Diseases)}                 & -              \\
				\midrule
				SapBERT \citep{Liu2021a}                                &                                                                 &                                                         &                \\
				\quad \cite{AbbreviationDeSohn2008} / KB / yes          & \cellcolor{blue!25}{93.2 (CTD Diseases) / 96.5 (CTD Chemicals)} & 92.3 (CTD Diseases)                                     & 50.4           \\
				\midrule
				BertOverkill \citep{Lai2021}                            &                                                                 &                                                         &                \\
				\quad \cite{AbbreviationDeSohn2008} / - / yes           & \cellcolor{blue!25}{93.2 (CTD Diseases) / 96.9 (CTD Chemicals)} & \cellcolor{red!25}{ 92.2 (CTD Diseases)}                & 55.0           \\
				\midrule
				LightweightBEL \citep{Chen2021}                         &                                                                 &                                                         &                \\
				\quad \cite{AbbreviationDeSohn2008} / - / no            & -                                                               & 89.59 (CTD Diseases)                                    & -              \\
				\midrule
				ClusteringInference \citep{Angell2021}                  &                                                                 &                                                         &                \\
				\quad \cite{AbbreviationDeSohn2008} / - / no            & 91.3 (UMLS) $\dagger$ $\ddag$                                   & -                                                       & 74.1   $\ddag$ \\
				\midrule
				MedWiki \citep{CrossDomainDaVarma2021}                  &                                                                 &                                                         &                \\
				\quad \cite{schwartz2002simple} / Wikipedia,PubMed / no & 91.5 (UMLS) $\dagger$ $\ddag$                                   & -                                                       & 74.6   $\ddag$ \\
				\midrule
				KRISSBERT \citep{zhang2022knowledge}                    &                                                                 &                                                         &                \\
				\quad - / PubMed / no                                   & 90.7 (UMLS) / 96.9 (UMLS) $\ddag$                               & 89.9 (UMLS)  $\ddag$                                    & 70.6  $\ddag$  \\
				\midrule
				GenBioEL \citep{GenerativeBiomYuan}                     &                                                                 &                                                         &                \\
				\quad \cite{AbbreviationDeSohn2008} / KB / yes          & 92.6 (MeSH) $\dagger$  $\diamondsuit$                           & \cellcolor{red!25}{ 91.6 (CTD Diseases)} $\diamondsuit$ & -              \\
				\midrule
				arboEL \citep{Agarwal2022}                              &                                                                 &                                                         &                \\
				\quad  - / - / no                                       & -                                                               & -                                                       & 72.31          \\
				\bottomrule
			\end{tabular}
		}
		\caption{Overview of experimental designs adopted by recent neural approaches for biomedical entity linking.
			We highlight with a color all cells reporting results
			which can be compared directly, i.e.: same corpus and knowledge base, abbreviation resolution (Abbr. Res.),
			pre-training (if any) and evaluation protocol.
			$\dagger$ No distinction between disease and chemical annotations
			$\ddag$ Reranking model
			$\diamondsuit$ Ablation study without pre-training
		}~\label{tab:approaches}
	\end{table*}

	Studies in BEL have converged to using primarily three corpora: \texttt{NCBI Disease}  \citep{Dogan2014},
	\texttt{BC5CDR} \citep{Li2016a} and \texttt{MedMentions} \citep{mohanmedmentions}.
	However, as shown in Table \ref{tab:approaches}, where we report a summary of recent approaches,
	experimental setups differ importantly in terms of the corpora and KBs used,
	making comparison based solely on published numbers problematic.
	For instance, in BioSyn the \textsc{BC5CDR} corpus is divided into two,
	distinguishing between chemical and disease entities, and linked to the CTD vocabularies,
	while GenBioEL reports results on the entire corpus linking to MeSH,
	preventig direct comparison.
	Notably, neural approaches rely on different pre-training strategies with different data sources,
	making the training signal vary significantly across approaches.
	This ultimately hinders estimating the difference in performance stemming purely from algorithmic differences.
	For instance MedWiki outperforms KRISSBERT on MedMentions,
	but as it is pre-trained on a larger pool of documents,
	it unclear whether this is due to differences in pre-training or model architecture.
	Similarly, it is not possible to directly estimate the impact of
	using abbreviation resolution in the reported performance.
	Finally, studies differ in the type of evaluation used,
	with some models deploying a lenient evaluation (see Section \ref{sec:syn_as_ent})
	further hindering direct comparison even if the corpus and the KB are the same.

	\section{Showcase}\label{sec:showcase}

	\begin{listing}[!htbp]
		\begin{minted}[xleftmargin=15pt,linenos]{python}
from collections import defaultdict(list)
from sqlalchemy import select
from belb import (AutoBelbKb, AutoBelbCorpus,
                  CORPUS_TO_KB, Tables)

from belb.resources import Corpora, Kbs

for corpus in Corpora:
  corpus = AutoBelbCorpus.from_name(corpus.name)

  kb = AutoBelbKb.from_name(CORPUS_TO_KB[corpus.name])

  table = kb.schema.get(Tables.KB)
  query = select(table.c.entity, table.c.name)

  name_to_entity = defaultdict(list)
  with kb as handle:
    for row in hadle.query(query):
      synonym = row["name"].lower()
      name_to_entity[synonym].append(row["entity"])

  for document in corpus["test"]:
    for a in document.annotations:
      name_to_entity.get(a.text.lower(), -1)
\end{minted}
		\caption{Example code to test an exact-match approach on BELB.}\label{lst:example_code}
	\end{listing}

	In Listing \ref{lst:example_code} we show that with BELB in less than 30 lines of code
	it is possible to test a simple exact-match approach on all its available pairs of corpus and knowledge base.

	\section{Corpora and Knowledge Bases}~\label{app:resources}

	\begin{table*}[!htbp]
		\centering
		% \resizebox{\textwidth}{!}{
		\begin{tabular}{l|l|l}
			                             & Website (link)                                                                                                                                                    & Lincense                    \\
			\toprule
			\textbf{Corpora}             &                                                                                                                                                                   &                             \\
			\quad \texttt{GNormPlus}     & \href{https://www.ncbi.nlm.nih.gov/research/bionlp/Tools/gnormplus/}{https://www.ncbi.nlm.nih.gov/research/bionlp/Tools/gnormplus/}                               & Public domain               \\
			\quad \texttt{NLM-Gene}      & \href{https://zenodo.org/record/5089049}{https://zenodo.org/record/5089049}                                                                                       & CC0 1.0                     \\
			\quad \texttt{NCBI Disease}  & \href{https://www.ncbi.nlm.nih.gov/CBBresearch/Dogan/DISEASE/}{https://www.ncbi.nlm.nih.gov/CBBresearch/Dogan/DISEASE/}                                           & Public domain               \\
			\quad \texttt{BC5CDR}        & \href{https://biocreative.bioinformatics.udel.edu/tasks/biocreative-v/track-3-cdr/}{https://biocreative.bioinformatics.udel.edu/tasks/biocreative-v/track-3-cdr/} & Public domain               \\
			\quad \texttt{NLM-Chem}      & \href{https://biocreative.bioinformatics.udel.edu/tasks/biocreative-vii/track-2/}{https://biocreative.bioinformatics.udel.edu/tasks/biocreative-vii/track-2/}     & CC0 1.0                     \\
			\quad \texttt{Linnaeus}      & \href{https://linnaeus.sourceforge.net/}{https://linnaeus.sourceforge.net/}                                                                                       & N/A                         \\
			\quad \texttt{S800}          & \href{https://species.jensenlab.org/}{https://species.jensenlab.org/}                                                                                             & N/A                         \\
			\quad \texttt{BioID}         & \href{https://biocreative.bioinformatics.udel.edu/tasks/biocreative-vi/track-1/}{https://biocreative.bioinformatics.udel.edu/tasks/biocreative-vi/track-1/}       & N/A                         \\
			\quad \texttt{MedMentions}   & \href{https://github.com/chanzuckerberg/MedMentions}{https://github.com/chanzuckerberg/MedMentions}                                                               & CC0 1.0                     \\
			\quad \texttt{SNP}           & \href{http://www.scai.fraunhofer.de/snp-normalization-corpus.html}{http://www.scai.fraunhofer.de/snp-normalization-corpus.html}                                   & N/A  $\ddag$                \\
			\quad \texttt{Osiris v1.2}   & \href{http://dx.doi.org/10.1186/1471-2105-9-84}{http://dx.doi.org/10.1186/1471-2105-9-84} $\dagger$                                                               & CC BY 3.0                   \\
			\quad \texttt{tmVar v3}      & \href{https://www.ncbi.nlm.nih.gov/research/bionlp/Tools/tmvar/}{https://www.ncbi.nlm.nih.gov/research/bionlp/Tools/tmvar/}                                       & Public domain               \\
			\midrule
			\textbf{Knowledge bases}     &                                                                                                                                                                   &                             \\
			\quad \textsc{CTD Diseases}  & \href{http://ctdbase.org/downloads/\#alldiseases}{http://ctdbase.org/downloads/\#alldiseases}                                                                     & Copyrighted  $\diamondsuit$ \\
			\quad \textsc{CTD Chemicals} & \href{https://ctdbase.org/downloads/\#allchems}{https://ctdbase.org/downloads/\#allchems}                                                                         & Copyrighted  $\diamondsuit$ \\
			\quad \textsc{NCBI Taxonomy} & \href{https://www.ncbi.nlm.nih.gov/taxonomy}{https://www.ncbi.nlm.nih.gov/taxonomy}                                                                               & Public domain               \\
			\quad \textsc{Cellosaurus}   & \href{https://www.cellosaurus.org/}{https://www.cellosaurus.org/}                                                                                                 & CC BY 4.0                   \\
			\quad \textsc{NCBI Gene}     & \href{https://www.ncbi.nlm.nih.gov/gene}{https://www.ncbi.nlm.nih.gov/gene}                                                                                       & Public domain               \\
			\quad \textsc{dbSNP}         & \href{https://www.ncbi.nlm.nih.gov/snp/}{https://www.ncbi.nlm.nih.gov/snp/}                                                                                       & Public domain               \\
			\quad \textsc{UMLS}          & \href{https://www.nlm.nih.gov/research/umls/index.html}{https://www.nlm.nih.gov/research/umls/index.html}                                                         & DUA                         \\
			\bottomrule
		\end{tabular}
		% }
		\caption{Corpora in BELB with corresponding link and license information.
			$\dagger$ The original website is no longer available: \href{https://sites.google.com/site/laurafurlongweb/databases-and-tools/corpora/}{https://sites.google.com/site/laurafurlongweb/databases-and-tools/corpora/}.
			The original corpus is redistributed by \citep{Thomas2016a}.
			$\ddag$ Commercial use is forbidden.
			$\diamondsuit$ Free for non-commercial uses, otherwise requires paid licensing
		}~\label{tab:license}
	\end{table*}

	In this section we describe in detail the corpora and knowledge bases
	contained in BELB, grouped by their entity type:

	\textbf{Gene} For genes, in contrast to previous approaches \citep{Tutubalina2020},
	we use \texttt{GNormPlus} \citep{Wei2015} instead of the
	BioCreative II Gene Normalization (BC2GN) corpus \citep{OverviewOfBioMorgan2008}.
	This is because BC2GN
	(i) is limited to human genes and
	(ii) it provides identifier annotations only at the \textit{document level}.
	\texttt{GNormPlus} consists of two corpora re-annotated at the \textit{mention level}, namely:
	BC2GN and the Gene Indexing Assistant (GIA) test collection\footnote{\url{https://ii.nlm.nih.gov/TestCollections/}}.
	We devote the GIA test collection as development split since GNormPlus (corpus) does not provide one.
	We include as well \texttt{NLM-Gene} \citep{Islamaj2021},
	which covers ambiguous gene names by including more species.
	As it offers only a train and a test split,
	we randomly sample 10\% (50 documents) of the training data to be used for development.
	Both corpora are linked to \textsc{NCBI Gene} \citep{Brown2015},
	which integrates detailed information for known and predicted genes,
	including data from all major taxonomic groups.

	\textbf{Disease} Standard corpora for disease normalization are \texttt{NCBI Disease} \citep{Dogan2014}
	and \texttt{BC5CDR} \citep{Li2016a}, originally created for the Chemical Disease Relation (CDR) track at BioCreative V.
	Entity mentions in these corpora are normalized to \textsc{CTD Diseases} \citep{ComparativeToxDavis2023},
	a.k.a. MEDIC, which is a modified subset of descriptors from the \enquote{Diseases} branch of MeSH \citep{lipscomb2000medical},
	combined with genetic disorders from the OMIM database \citep{hamosh2005online}.

	\textbf{Chemical} For chemicals we use \texttt{BC5CDR} as it provides as well chemical annotations
	and additionally include
	\texttt{NLM-Chem}, a collection of \textit{full text} articles.
	Specifically we employ the version released for the NLM-Chem track
	at BioCreative VII \citep{NlmChemBc7MIslama2022}.
	Mentions in these corpora are linked to the chemical branch of MeSH,
	which however includes broad categorises such as \enquote{D014867} (\enquote{water}).
	We therefore only retain entity mentions linked to \textsc{CTD Chemicals} \citep{ComparativeToxDavis2023},
	which merges descriptors from the \enquote{Chemicals and Drugs} category and Supplementary Concept Records in MeSH
	and removes several branches of the original MeSH hierarchy
	if either (i) are not molecular reagents/clinical drugs, e.g.\enquote{Purines},
	or (ii) are broad chemical classes, e.g. \enquote{Poisons}.
	% a redacted version of MeSH specialized for chemicals.

	\textbf{Species} For species normalization, we use the \texttt{Linnaeus} corpus \citep{LinnaeusASpeMartin},
	a collection of annotated full text articles, and \texttt{S800} \citep{Pafilis2013},
	containing 100 abstracts from 8 different publication categories (e.g. bacteriology).
	Both corpora are linked to \textsc{NCBI Taxonomy} \citep{TheNcbiTaxonoScott2012},
	the standard nomenclature and classification repository of organisms
	for sequences databases such as GenBank.
	Both corpora do not provide canonical splits.
	We rely on those defined in \url{https://github.com/spyysalo/linnaeus-corpus/tree/master/split}
	and \url{https://github.com/spyysalo/s800} for \texttt{Linnaeus} and \texttt{S800} respectively.

	\textbf{Cell line} To the best of our knowledge, the only available corpus
	offering normalized entity mentions of cell lines is \texttt{BioID}  \citep{arighi2017bio},
	introduced in the Interactive Bio-ID Assignment Track at BioCreative VI.
	In contrast to all other corpora, text in BioID are figure captions found in full text articles.
	The corpus is annotated with mentions of multiple entity types (e.g. chemicals and species),
	but we retain exclusively those linked to \textsc{Cellosaurus} \citep{TheCellosaurusBairoc2018},
	a nomenclature of cell lines used in biomedical research.
	Only the annotated training split was made available for the BioCreative track .
	We therefore re-structure the training data into 80/10/10 train, development and test split respectively.

	\textbf{Variant} The de facto standard corpora for variants are:
	\texttt{SNP} \citep{ChallengesInTThomas2011},
	\texttt{Osiris v1.2} \citep{Osirisv12ANFurlon2008} and
	\texttt{tmVar v3} \citep{Wei2022}.
	All corpora contain exclusively a test split.
	The \texttt{tmVar (v3)} corpus offers multiple normalizations for a given entity mention,
	e.g. a CAR Canonical Allele Identifier (CA ID) or its corresponding gene.
	For all corpora, we keep only the entity mentions which are normalized to \textsc{dbSNP},
	the primary knowledge base for human genetic variants.
	Despite its name, dbSNP contains a wide range of polymorphisms besides Short Nucleotide Polymorphisms, e.g.
	short deletion and insertion polymorphisms and multinucleotide polymorphisms.

	\textbf{UMLS} For comparison with previous studies we include
	the \texttt{MedMentions} corpus with the 2017AA full release of \textsc{UMLS}.
	We note that although \textsc{CTD Diseases} and \textsc{CTD Chemicals} are technically a subset of UMLS,
	we keep them as separate KBs because (i) some corpora were specifically linked to them (e.g. \texttt{NCBI Disease}) and
	(ii) with UMLS containing many unrelated concepts, it would unnecessarily increase the search space.

	In Table \ref{tab:license} we report links to all resources included in BELB along with the license with which they are released.
	All resources created by NLM, as a governmental agency,
	are by nature of public domain\footnote{See \url{https://www.ncbi.nlm.nih.gov/CBBresearch/Dogan/DISEASE/disclaimer.html} for the public domain notice which usually accompanies these resources.}.
	Many corpora do not specify a license (we mark them as N/A) but can be freely accessed
	and do not specify any limitation on the data usage (except in forbidding commercial use).
	The same considerations hold for the knowledge bases.
	UMLS is the only resource which is not freely available and users are required to enter a Data Usage Agreement (DUA).

	\section{MedMentions}\label{sec:medmentions}

	\begin{table}[!h]
		\centering
		\resizebox{\columnwidth}{!}{
			\begin{tabular}{l|l|l}
				   & Semantic Group               & Entity mentions (\%) \\
				\toprule
				1  & Phenomena                    & 41,422  (20\%)       \\
				2  & Chemicals \& Drugs           & 37,401  (18\%)       \\
				3  & Procedures                   & 33,478  (16\%)       \\
				4  & Concepts \& Ideas            & 22,824  (11\%)       \\
				5  & Anatomy                      & 22,248  (11\%)       \\
				6  & Living Beings                & 19,626  (10\%)       \\
				7  & Disorders                    & 18,081  (0.9\%)      \\
				8  & Organizations                & 2,143   (0.1\%)      \\
				9  & Devices                      & 2,018   (0.1\%)      \\
				10 & Physiology                   & 1,773   (0.1\%)      \\
				11 & Objects                      & 1,352   (0.1\%)      \\
				12 & Occupations                  & 9,16    (0\%)        \\
				13 & Activities \& Behaviors      & 0      (0\%)         \\
				14 & Genes \& Molecular Sequences & 0      (0\%)         \\
				15 & Geographic Areas             & 0      (0\%)         \\
				\bottomrule
			\end{tabular}
		}
		\caption{Ranked number of entity mentions (and relative amount) in MedMentions (ST21PV) grouped by UMLS Semantic Groups.}~\label{tab:medmention}
	\end{table}

	In Table \ref{tab:medmention} we report the ranked number of entity mentions in \textsc{MedMentions} (ST21PV)
	grouped by their UMLS Semantic Groups\footnote{\url{https://lhncbc.nlm.nih.gov/ii/tools/MetaMap/documentation/SemanticTypesAndGroups.html}}.
	As reported in Section \ref{sec:analysis:umls} entity types such as diseases (\enquote{Disorders})
	and species (\enquote{Living Beings}) are covered only marginally, while genes (\enquote{Genes \& Molecular Sequences})
	are completely absent. Notably, a high proportion of mentions ($\sim$47\%)
	is devoted instead to general entities such as \enquote{Phenomena} and \enquote{Procedures}.

	\section{Unified schema for knowledge bases}\label{app:schema}

	\begin{listing*}[!htbp]
		\begin{minted}{sql}
CREATE TABLE kb(
    uid INTEGER PRIMARY KEY,
    entity INTEGER NOT NULL,
    description INTEGER NOT NULL,
    name TEXT NOT NULL,
    foreign_entity INTEGER REFERENCES foreign_entities(entity),
)

CREATE TABLE foreign_entities(
    entity INTEGER NOT NULL,
    name TEXT NOT NULL,
)

CREATE TABLE history(
    old INTEGER PRIMARY KEY,
    new INTEGER NOT NULL,
)
\end{minted}
		\caption{Example SQL code defining the BELB schema for all knowledge bases.}\label{lst:schema}
	\end{listing*}

	In Listing \ref{lst:schema} we provide an overview of the unified schema
	used to store all knowledge bases in BELB.
	In its basic form, a KB is a list of synonyms (names),
	each associated with a single entity.
	All KBs provide as well information about each name (\enquote{description}).
	For instance in \textsc{NCBI Taxonomy} a name can be the \enquote{scientific name} of a species (\enquote{Homo Sapiens})
	or the \enquote{common} one  (\enquote{human}).
	In all KBs exclusively one name must be the primary one,
	i.e. the one most commonly used to refer to the concept represented by the entity.
	% We call this the \textit{symbol} (in UMLS this is called the \enquote{Preferred Name}).
	For instance, the primary name (a.k.a symbol) for entity \enquote{2} in \textsc{NCBI Gene} is \enquote{A2M}.
	Some of the KBs are interconnected.
	For instance, \enquote{$\alpha$2microglobulin} represents a different entity when referring to the human or rat gene.
	Thus, besides having different identifiers (\enquote{2} and \enquote{24153} respectively),
	to ease downstream applications (e.g. data integration),
	\textsc{NCBI Gene} provides what we call (borrowing from the database jargon) a \textit{foreign} entity.
	% To handle homonyms, some KBs require information from a \textit{foreign KB}
	% to fully characterize their entries, specifically the identifiers (\textit{foreign identifiers}).
	For instance, all entries in NCBI Gene with identifier 2 are accompanied by the foreign entity \enquote{9606},
	i.e. the entity in \textsc{NCBI Taxonomy} denoting \enquote{Homo Sapiens} (human).
	If available, a KB provides as well a \textit{history} table, where changes to the identifiers are tracked,
	i.e. if they have been replaced by others or have become obsolete.

	\section{Updated corpora}\label{app:update_corpora}

	\begin{table}[!h]
		\centering
		\resizebox{6cm}{!}{
			\begin{tabular}{l|c|c|l}
				                      & \multicolumn{3}{c}{Entities}                          \\
				                      & Replaced                     & Removed & Total        \\
				\hline
				\hline
				\texttt{GNormPlus}    & -                            & 1       & 1 (0.03\%)   \\
				\texttt{NLM-Gene}     & 1                            & 1       & 2 (0.07\%)   \\
				\texttt{NCBI Disease} & -                            & 11      & 11 (0.25\%)  \\
				\texttt{BC5CDR (C)}   & -                            & 243     & 243 (4.56\%) \\
				\texttt{NLM-Chem}     & -                            & 36      & 36 (0.31\%)  \\
				\texttt{Linnaeus}     & 51                           & -       & 51 (3.57\%)  \\
				\texttt{S800}         & 2                            & -       & 2 (0.26\%)   \\
				\texttt{BioID}        & -                            & -       & 0 (0.0\%)    \\
				\texttt{SNP}          & 6                            & 5       & 11 (2.13\%)  \\
				\texttt{Osiris v1.2}  & 7                            & -       & 7 (2.68\%)   \\
				\texttt{tmVar v3}     & 5                            & -       & 5 (0.49\%)   \\
				\texttt{MedMentions}  & -                            & -       & 0 (0.0\%)    \\
			\end{tabular}
		}
		\caption{Overview of the number of changes in entity labels in the test split of each corpus in BELB. }~\label{tab:corpora:update}
	\end{table}

	In Table \ref{tab:corpora:update} we report the number of changes in entity label in the test split of each BELB corpus.
	For each gold label associated to an entity mention there can be two types of changes.
	Either the label has been replaced (Replaced), and in this case it can be updated,
	or it was removed from the KB (Removed), which makes the label obsolete and the mention not linkable,
	in which case we exclude the entity mention from the test set.

	\section{Biomedical entity liking systems}~\label{app:systems}

	\begin{table*}[!htbp]
		\centering
		% \resizebox{\textwidth}{!}{
		\begin{tabular}{l|l}
			                                    & Implementation (link)                                                                                                               \\
			\toprule
			\textbf{Rule-based entity-specific} &                                                                                                                                     \\
			\quad GNormPlus                     & \href{https://www.ncbi.nlm.nih.gov/research/bionlp/Tools/gnormplus/}{https://www.ncbi.nlm.nih.gov/research/bionlp/Tools/gnormplus/} \\
			\quad TaggerOne                     & \href{https://www.ncbi.nlm.nih.gov/research/bionlp/Tools/taggerone/}{https://www.ncbi.nlm.nih.gov/research/bionlp/Tools/taggerone/} \\
			\quad tmVar v3                      & \href{https://ftp.ncbi.nlm.nih.gov/pub/lu/tmVar3/}{https://ftp.ncbi.nlm.nih.gov/pub/lu/tmVar3/}                                     \\
			\quad BC7T2                         & \href{https://github.com/bioinformatics-ua/biocreativeVII\_track2}{https://github.com/bioinformatics-ua/biocreativeVII\_track2}     \\
			\quad SR4GN                         & \href{https://www.ncbi.nlm.nih.gov/research/bionlp/Tools/sr4gn/}{https://www.ncbi.nlm.nih.gov/research/bionlp/Tools/sr4gn/}         \\
			\quad FuzzySearch                   & \href{https://github.com/maxbachmann/RapidFuzz}{https://github.com/maxbachmann/RapidFuzz}                                           \\
			\midrule
			\textbf{Pre-trained language model} &                                                                                                                                     \\
			\quad arboEL                        & \href{https://github.com/dhdhagar/arboEL}{https://github.com/dhdhagar/arboEL}                                                       \\
			\quad GenBioEl                      & \href{https://github.com/Yuanhy1997/GenBioEL}{https://github.com/Yuanhy1997/GenBioEL}                                               \\
			\quad BioSyn                        & \href{https://github.com/dmis-lab/BioSyn}{https://github.com/dmis-lab/BioSyn}                                                       \\
			\bottomrule
		\end{tabular}
		% }
		\caption{Overview of the biomedical entity linking systems benchmarked on BELB.}~\label{tab:systems:overview}
	\end{table*}

	In Table \ref{tab:systems:overview} we report all systems (and the link to their original implementation)
	taken into consideration in our benchmarking.

	\subsection{Rule-based entity-specific systems}\label{app:systems:rb}

	\begin{table*}[!htbp]
		\centering
		% \resizebox{\textwidth}{!}{
		\begin{tabular}{l|l|l|l}
			            & Entity type(s)     & Recognition       & Linking                            \\
			\toprule
			GNormPlus   & Gene               & CRF               & Inference Network + TF-IDF + Rules \\
			TaggerOne   & Disease, Cell line & semi-Markov model & TF-IDF                             \\
			BC7T2W      & Chemical           & -                 & String matching + Embeddings       \\
			SR4GN       & Species            & String matching   & String matching + Rules            \\
			tmVar       & Variant            & String matching   & String matching + Rules            \\
			SciSpacy    & UMLS               & -                 & TF-IDF                             \\
			FuzzySearch & -                  & -                 & Levenshtein Distance               \\
			\bottomrule
		\end{tabular}
		% }
		\caption{Overview of the rule-based entity-specific baselines benchmarked on BELB.}~\label{tab:systems:rb}
	\end{table*}

	In Table \ref{tab:systems:rb} we provide an overview of the rule-based entity-specific models.
	Implementation details for each model are reported below.

	\textbf{GNormPlus}  uses a CRF Model to perform entity recognition.
	The normalization component is a statistical inference network based on TF-IDF frequencies.
	The system comes with two pre-trained models, namely:
	\enquote{GNR.Model}, which was trained on the train and development split (as defined by BELB) of \texttt{GNormPlus}
	iand \enquote{GNR.GNormPlusCorpus\_\\NLMGeneTrain.Model},
	which was trained on the whole \texttt{GNormPlus} corpus and the train and development split
	(as defined by BELB) of the \texttt{NLM-Gene} corpus.
	We use the first one when evaluating on \texttt{GNormPlus} and the second on \texttt{NLM-Gene}.

	\noindent \textbf{SR4GN} (Species Recognition for Gene Recognition) is a rule-based system which, as the name suggests,
	is mainly a support component for gene normalization.
	It implements several custom rules to address cases where species information is not explicitly available.
	We run GNormPlus as well on the species corpora as SR4GN is only available as a GNormPlus component.

	\noindent \textbf{tmVar} uses pattern matching to both recognize and normalize variants,
	falling back to dictionary-lookup when the first fails.
	It is distributed with a pre-trained model and a pre-processed version of dbSNP.
	Its usage depends on GNormPlus, as it requires normalized gene mentions to perform linking.

	\noindent \textbf{TaggerOne} is a general purpose joint recognition and linking system based on a semi-Markov model.
	It provides two models: \enquote{model\_NCBID.bin} and \enquote{model\_BC5CDRD.bin}
	which are trained on the train split of \texttt{NCBI Disease} and the disease annotations of \texttt{BC5CDR} respectively.
	\citet{Wei2019a} reports to use a TaggerOne model trained on \texttt{BioID}.
	They however do not publicly release the trained model. Our attempt to use TaggerOne training code resulted in a error.

	% Although it is trainable and not entity-specific, we list it in this section as its normalization component is based on TF-IDF vectors.
	% In PubTator TaggerOne is used as well for the cell line entity type, however the corresponding model was never publicly released.
	% We tried training the system on the \texttt{BioID} corpus but encountered a runtime error.
	% We hypothesize that the authors used different unpublished code to train the cell line model.

	\noindent \textbf{BC7T2W} is a hybrid model based on dictionary-lookup and BioBERT embeddings \citep{lee2020biobert}.
	By default the systems includes in the target KB all mentions in \texttt{NLM-Chem} with their gold labels.
	We modify the \enquote{settings.yaml} file to disable this option.

	\noindent \textbf{SciSpacy} comes with a joint entity recognition and linking pipeline.
	However, it is possible to run only the linking component
	using the abbreviation detection module separately
	(we use the one integrated in the \enquote{en\_core\_sci\_sm} model\footnote{\url{https://s3-us-west-2.amazonaws.com/ai2-s2-scispacy/releases/v0.5.1/en_core_sci_sm-0.5.1.tar.gz}})
	and feeding the result to linking system (\enquote{scispacy\_linker}).

	\noindent \textbf{FuzzySearch} This approach assigns a score in $[0,1]$ to pairs of strings.
	The score is defined as $\frac{D}{\vert s_{i} \vert  + \vert s_{i} \vert }$,
	where $D$ is the edit distance between strings $s_{i}$ and $s_{j}$
	and $\vert s_{i} \vert$ is the number of characters in $s_{i}$.
	We use the implementation provided by \url{https://github.com/maxbachmann/RapidFuzz}.
	Given a test mention we compute the similarity score with all synonyms in a KB
	and select the one with the highest score.

	\subsection{Pre-trained language models}~\label{app:systems:plm}

	We convert all corpora and KBs in BELB in the format required by each model and re-train them with the default hyper-parameters
	reported by the original implementations. All models are trained and evaluated using two Nvidia Tesla A100.

	\noindent \textbf{BioSyn} We use the default hyper-parameters provided by the authors.
	Unlike the original approach, we exclusively train on the train split of corpora.
	The backbone pre-trained language model is BioBERT.

	\noindent \textbf{GenBioEl} is initialized with BART (\enquote{large}) weights \citep{BartDenoisingLewis2020}.
	It uses different values for learning rate and warmup steps for NCBI Disease and BC5CDR.
	We cannot perform a full hyper-parameter search for each corpus,
	and therefore select the values that work best for both corpora,
	i.e. a learning reate of $1e-5$ and 500 warmup steps.

	\noindent \textbf{arboEL} is initialized with BioBERT.
	The inference procedure for arboEL is parametrized by
	the number of $k$ nearest neighbor used to construct the graph (determining which pairs of nodes are connected).
	The implementation provided by the authors runs the inference trying different $k \in \{0, 1, 2, 4, 8\}$.
	For fair comparison with other models, we do not perform any hyperparameter optimization
	and hence report the score for $k=0$.

	\section{Additional results}

	\begin{table}[!h]
		\centering
		\begin{tabular}{l|lll}
			                &                       & Corpora              &                   \\
			\toprule
			                & \texttt{GNormPlus}    & \texttt{NLM-Gene}    &                   \\
			\quad GNormPlus & 0.93                  & 0.73                 &                   \\
			\hline
			                & \texttt{Linnaeus}     & \texttt{S800}        &                   \\
			\quad SR4GN     & 0.80                  & 0.70                 &                   \\
			\hline
			                & \texttt{NCBI Disease} & \texttt{BC5CDR (D)}  &                   \\
			\quad TaggerOne & 0.82                  & 0.82                 &                   \\
			\hline
			                & \texttt{SNP}          & \texttt{Osiris v1.2} & \texttt{tmVar v3} \\
			\quad tmVar     & 0.91                  & 0.67                 & 0.92              \\
			\bottomrule
		\end{tabular}
		\caption{NER recall of rule-based entity-specific systems which perform joint recognition and normalization.}~\label{tab:ner_recall}
	\end{table}

	\begin{table}[!h]
		\centering
		\begin{tabular}{l|l|l|l|l}
			                            & RBES & BioSyn & GenBioEl & arboEL \\
			\toprule

			\textbf{Disease}            & 0.94 & 0.84   & 0.89     & 0.87   \\
			\quad \texttt{NCBI Disease} & 0.94 & 0.82   & 0.87     & 0.86   \\
			\quad \texttt{BC5CDR (D)}   & 0.94 & 0.87   & 0.92     & 0.88   \\
			\hline
			\textbf{Chemical}           & 0.72 & 0.72   & 0.81     & 0.77   \\
			\quad \texttt{BC5CDR (C)}   & 0.82 & 0.85   & 0.95     & 0.88   \\
			\quad \texttt{NLM-Chem}     & 0.67 & 0.67   & 0.75     & 0.72   \\
			\hline
			\textbf{Cell line}          &      &        &                   \\
			\quad \texttt{BioID}        & 0.80 & 0.81   & 0.95     & 0.95   \\
			\hline
			\textbf{Species}            & 0.97 & 0.91   & 0.85     & 0.76   \\
			\quad \texttt{Linnaeus}     & 0.99 & 0.93   & 0.81     & 0.74   \\
			\quad \texttt{S800}         & 0.93 & 0.88   & 0.93     & 0.79   \\
			\hline
			\textbf{Gene}               & 0.81 & -      & 0.05     & 0.30   \\
			\quad \texttt{GNormPlus}    & 0.86 & OOM    & 0.06     & 0.36   \\
			\quad \texttt{NLM-Gene}     & 0.75 & OOM    & 0.03     & 0.23   \\
			\hline
			\textbf{Variant}            & 0.90 &        &                   \\
			\quad \texttt{SNP}          & 0.94 & -      & -        & -      \\
			\quad \texttt{Osiris v1.2}  & 0.91 & -      & -        & -      \\
			\quad \texttt{tmVar v3}     & 0.88 & -      & -        & -      \\
			\midrule
			\textbf{UMLS}               &      &        &                   \\
			\quad \texttt{MedMentions}  & 0.58 & OOM    & 0.41     & 0.69   \\
			\bottomrule
		\end{tabular}
		\caption{Performance of all baselines on BELB (test set) with strict evaluation (see Section \ref{sec:syn_as_ent}).
			All scores are \textit{mention-level} recall@1.
			OOM: out-of-memory ($>$200GB)
		}~\label{tab:results:mention_strict}
	\end{table}

	\begin{table}[!h]
		\centering
		% \resizebox{\textwidth}{!}{
		\begin{tabular}{l|l|l|l|l}
			                            & RBES & BioSyn & GenBioEL & arboEL \\
			\toprule
			\textbf{Disease}            &      & 0.84   & 0.88     & 0.84   \\
			\quad \texttt{NCBI Disease} & -    & 0.81   & 0.84     & 0.80   \\
			\quad \texttt{BC5CDR (D)}   & -    & 0.85   & 0.89     & 0.85   \\
			\hline
			\textbf{Chemical}           & 0.72 & 0.72   & 0.81     & 0.77   \\
			\quad \texttt{BC5CDR (C)}   & 0.82 & 0.85   & 0.95     & 0.87   \\
			\quad \texttt{NLM-Chem}     & 0.67 & 0.67   & 0.75     & 0.72   \\
			\hline
			\textbf{Cell line}          &      &        &          &        \\
			\quad \texttt{BioID}        & 0.82 & 0.82   & 0.96     & 0.95   \\
			\hline
			\textbf{Species}            &      & 0.87   & 0.81     & 0.76   \\
			\quad \texttt{Linnaeus}     & -    & 0.89   & 0.77     & 0.75   \\
			\quad \texttt{S800}         & -    & 0.83   & 0.89     & 0.79   \\
			\hline
			\textbf{Gene}               &      & -      & 0.18     & 0.28   \\
			\quad \texttt{GNormPlus}    & -    & OOM    & 0.20     & 0.35   \\
			\quad \texttt{NLM-Gene}     & -    & OOM    & 0.10     & 0.21   \\
			\hline
			\textbf{Variant}            &      &        &          &        \\
			\quad \texttt{SNP}          & -    & -      & -        & -      \\
			\quad \texttt{Osiris v1.2}  & -    & -      & -        & -      \\
			\quad \texttt{tmVar v3}     & -    & -      & -        & -      \\
			\midrule
			\textbf{UMLS}               &      &        &          &        \\
			\quad \texttt{MedMentions}  & 0.58 & OOM    & 0.57     & 0.69   \\
			\bottomrule
		\end{tabular}
		% }
		\caption{Performance of all baselines on the full test set of all BELB corpora (see Section \ref{sec:disentangling_ner_nen}).
			All scores are \textit{mention-level} recall@1. OOM: out-of-memory ($>$200GB)
		}~\label{tab:results:full}
	\end{table}

\end{appendices}

\end{document}